\documentclass[pdflatex,sn-standardnature]{sn-jnl}
\jyear{2025}
\usepackage{setspace}
\usepackage{lineno}
\usepackage{graphicx}
\usepackage{multirow}
\usepackage{amsmath,amssymb,amsfonts}
\usepackage{amsthm}
\usepackage{mathrsfs}
\usepackage[title]{appendix}
\usepackage{xcolor}
\usepackage{textcomp}
\usepackage{manyfoot}
\usepackage{booktabs}
\usepackage{algorithm}
\usepackage{algorithmicx}
\usepackage{algpseudocode}
\usepackage{listings}
\usepackage{textcomp}
\usepackage{array}
\usepackage{threeparttable}
\usepackage{float}
\usepackage{caption}
\usepackage{multirow}
\usepackage{makecell}
\usepackage{adjustbox}
\usepackage{paralist}
\usepackage{fancybox}

\renewcommand{\figurename}{Fig.}
\renewcommand{\tablename}{Tab.}

\begin{document}

%% Title
\title[Article Title]{ClimaEmpact: Domain-Aligned Small Language Models and Datasets for Extreme Weather Analytics}

%% Authors
\author[1]{\fnm{Deeksha} \sur{Varshney}}\email{}
\author[1]{\fnm{Keane} \sur{Ong}}\email{}
\author[3]{\fnm{Rui} \sur{Mao}}\email{}
\author[3]{\fnm{Erik} \sur{Cambria}}\email{}
\author*[2]{\fnm{Gianmarco} \sur{Mengaldo}}\email{mpegim@nus.edu.sg}

% Affiliations
\affil[1]{\orgname{College of Design and Engineering, National University of Singapore}, \\ \country{Singapore}}

\affil[2]{\orgname{Department of Mathematics, National University of Singapore}, \\ \country{Singapore}}

\affil[3]{\orgname{College of Computing and Data Science, Nanyang Technological University}, \\ \country{Singapore}}

% \affiliation[label1]{organization={National University of Singapore}, 
% city={Singapore}, 
% country={Singapore}}

% \affiliation[label2]{organization={Nanyang Technological University}, 
% city={Singapore}, 
% country={Singapore}}

%TC:ignore
% \abstract{
% }

\abstract{
Accurate assessments of extreme weather events are vital for research and policy, yet localized and granular data remain scarce in many parts of the world. 
This data gap limits our ability to analyze potential outcomes and implications of extreme weather events, hindering effective decision-making. 
Large Language Models (LLMs) can process vast amounts of unstructured text data, extract meaningful insights, and generate detailed assessments by synthesizing information from multiple sources. 
Furthermore, LLMs can seamlessly transfer their general language understanding to smaller models, enabling these models to retain key knowledge while being fine-tuned for specific tasks.
In this paper, we propose \textit{Extreme Weather Reasoning-Aware Alignment (EWRA)}, a method that enhances small language models (SLMs) by incorporating structured reasoning paths derived from LLMs, and ExtremeWeatherNews, a large dataset of extreme weather event-related news articles. 
EWRA and ExtremeWeatherNews together form the overall framework, ClimaEmpact, that focuses on addressing three critical extreme-weather tasks: categorization of tangible vulnerabilities/impacts, topic labeling, and emotion analysis. 
By aligning SLMs with advanced reasoning strategies on ExtremeWeatherNews (and its derived dataset ExtremeAlign used specifically for SLM alignment), EWRA improves the SLMs' ability to generate well-grounded and domain-specific responses for extreme weather analytics. 
Our results show that the approach proposed guides SLMs to output domain-aligned responses, surpassing the performance of task-specific models and offering enhanced real-world applicability for extreme weather analytics.
}

\keywords{Extreme weather, Large Language Models (LLMs), Small Language Models (SLMs), Alignment, Reasoning}

\maketitle
%TC:endignore

%%%%%%%%%%%%%%%%%%%%%%%%%%%%
% 1. INTRODUCTION
%%%%%%%%%%%%%%%%%%%%%%%%%%%%

\section{Introduction}
The growing occurrence and intensity of extreme weather events present significant dangers that affect both individuals and communities~\citep{masson2021ipcc}. 
These events, including heatwaves, prolonged dry conditions, extreme precipitation, and tropical cyclones, have significant effects on human livelihoods and the natural environment, often leading to lasting and sometimes irreversible consequences. 
However, our limited understanding of their societal impacts hinders the development of effective disaster response and communication strategies~\citep{mao2024understanding,camps2025artificial}. 
Enhancing access to comprehensive and accurate information is crucial for addressing this challenge. 
Such data can serve as a foundation for extreme weather analysis tasks that can provide crucial \textit{extreme weather analytics} such as vulnerability and impact assessment~\citep{hammond2015urban,zennaro2021exploring,coletti2013support,bechtel2023ready,eriksen2007developing}.

Large Language Models (LLMs) present a promising tool to provide comprehensive and accurate information on extreme weather events by efficiently extracting and synthesizing big climate-related data from diverse textual sources~\citep{li2024using,bulian2024assessing,wang2024exploring,zhu2023climate}.
Yet, there are several challenges on the pathway to achieve this goal, including outdated data that does not consider the latest events~\citep{zheng2024large}, and scarcity of expert-annotated data, which is critical for training and evaluation.  
These challenges can affect the accuracy of extreme weather analysis tasks by hindering the model's ability to account for the latest extreme weather events, shifts in vulnerability, and evolving impact assessments. 

To mitigate these issues, some solutions have been proposed in the literature. For instance, ClimateGPT~\citep{thulke2024climategpt} implements various strategies, including a hierarchical retrieval augmentation strategy, drawing upon external knowledge sources like Wikipedia and IPCC AR6 reports.  
While integrating scientifically validated sources enhances the accuracy of LLM responses, it does not entirely eliminate the potential for generating inaccurate or misleading information~\citep{vaghefi2023chatclimate}.

Indeed, the critical gap posed by the lack of structured and up-to-date knowledge in the domain of extreme weather event analysis still persists, and requires urgent attention.
Addressing this gap is crucial to comprehensively understand the evolving vulnerabilities and impacts associated with extreme weather events, encompassing both the tangible damages and the emotional consequences experienced by affected communities.

To fill this gap, we leverage the advanced reasoning capabilities of LLMs on news articles associated with extreme weather events. 
In particular, we prompt LLMs to generate detailed, step-by-step reasoning paths on three extreme-weather analysis tasks: (i) identification of vulnerability, impact, and emergency response of the region affected by the extreme event, (ii) topic/subtopic labeling and keyword extraction, and (iii) emotion analysis.  
These reasoning paths act as alignment data, providing clear, detailed explanations that illustrate how to approach and solve complex extreme weather analysis problems. 
For instance, when tasked with categorizing a sentence that describes a weather impact, the LLM does not simply provide the category but also explains the logical steps it followed, highlighting relevant keywords and contextual clues.

Building upon these alignment examples, we propose a novel domain-specific reasoning approach aimed at enhancing the analytical capabilities of language models for extreme weather analysis tasks. 
We introduce Extreme Weather Reasoning-Aware Alignment (EWRA), a fine-tuning method that transfers the advanced reasoning skills of LLMs to small language models (SLMs). 
EWRA trains SLMs using LLM-generated reasoning paths, enabling these models to decompose complex problems into manageable steps and provide clear, logical explanations, making them domain-specialized for providing extreme weather analytics. 
Our approach effectively applies a two-stage fine-tuning strategy: first, SLMs are trained with implicit reasoning to internalize reasoning logic instead of depending only on prompt patterns; second, the models are provided with detailed definitions of task categories within the prompts for explicit fine-tuning. 
This capability is particularly critical for retrieving useful extreme weather analytics, where robust representations capable of accurately identifying and characterizing extreme weather features and understanding the underlying rationale for predictions are as important as the predictions themselves~\citep{turbe2023evaluation,mengaldo2025progress}.

As part of this work, we also release a new dataset, ExtremeWeatherNews, which comprises news articles collected for 60 distinct extreme weather events. 
EWRA and ExtremeWeatherNews constitute the two elements of our ClimaEmpact framework (\url{climaempact.ai}) for near real-time extreme weather analytics. 
To evaluate the effectiveness of EWRA, we perform a series of experiments comparing its performance against baseline models on a specialized dataset derived from ExtremeWeatherNews, namely the ExtremeAlign dataset, prepared using LLM-generated reasoning paths. 
We evaluate multiple fine-tuning strategies, including standard and reasoning-based approaches, and show that EWRA delivers competitive performance compared to other fine-tuning methods across tasks, particularly on the Qwen2.5-3B-Instruct model. On the Vulnerability/Impact/Emergency assessment task, EWRA achieves 5.2\% improvements in Spearman Rank Correlation over ReasonExplicit-SFT, demonstrating stronger alignment with human-annotated reasoning patterns, while also showing comparable or superior performance on topic/subtopic labeling tasks and emotion analysis.

In summary, the key contributions of our work are as follows.
\begin{enumerate}
    \item We introduce Extreme Weather  Reasoning-Aware Alignment (EWRA), a novel approach for enhancing SLMs by transferring reasoning capabilities from LLMs through fine-tuning using synthetic alignment data for extreme weather event analysis.
    \item We release ExtremeWeatherNews, a comprehensive collection of news articles related to 60 distinct extreme weather events, curated to support the study of various extreme weather related tasks. In order to implement EWRA, we also introduce a new alignment dataset called the ExtremeAlign, comprising rationales across three tasks.
    \item We evaluate the performance of EWRA through a series of experiments using the ExtremeAlign dataset, demonstrating its ability to significantly improve domain-specific reasoning accuracy, as compared to task-specific baseline models.
\end{enumerate}
The ClimaEmpact framework is deployed online for fast and near real-time analyses of extreme weather events (Appendix \ref{sec:online_dash}). 

\section{Related Work}
\label{sec:related-work}

We focus on three key areas of research that are related to our work, namely: standard NLP methods, LLMs, and SLMs alignment. 
Each of these areas and the related work is reported in the following. \\

\noindent {\textbf{Traditional NLP methods for extreme weather.}}
Textual data, such as news reports, social media posts, and official bulletins, complement satellite-based datasets by providing context-rich and timely information on extreme weather impacts. 
Multiple natural language processing (NLP) approaches have utilized textual data to analyze and classify extreme weather impacts~\citep{de2024uncovering, sodoge2024text, alencar2024flash}. Social media data have also been exploited; for example, drought impacts in California have been categorized using BERT-based models applied to Twitter posts~\citep{zhang2022tweetdrought}. Similarly, automated text processing has mapped trajectories, impacts, and the aftermath of the 2021 European floods, providing broader categorization and insights into weather extremes~\citep{kahle2022classifying}.

Transformer-based models, such as ClimateBERT, have enhanced various NLP applications related to climate, including text classification, sentiment analysis, and fact-checking~\citep{webersinke2021climatebert}. Moreover, affective computing and text mining techniques have proven valuable for capturing emotional nuances within textual datasets~\citep{cambria2024senticnet,7pillars}. An example includes analyzing public opinions on wildfires through a neurosymbolic approach employing BERT-based models~\citep{duong2024neurosymbolicwildfire}. Additionally, the ClimaText dataset has demonstrated the limitations of keyword-based approaches in climate discourse, emphasizing the advantages of context-aware methods like BERT for effective topic labeling~\citep{varini2020climatext, chen2023ontology}. Recent studies have explored metaphorical concept mappings to analyze public perceptions of weather disasters, leveraging tools such as MetaPro to elucidate distinct public perspectives across various disaster types~\citep{mao2024understanding, mao2023metaproonline}.\\

\noindent \textbf{LLMs for extreme weather.}
Current methods typically rely on task-specific models, limiting adaptability and efficiency. Large language models (LLMs) offer an integrated framework, leveraging unsupervised pretraining, instruction tuning, and reinforcement learning to enhance contextual understanding from extensive textual datasets~\citep{zhou2023lima}. These capabilities are increasingly utilized for disaster response tasks. Examples include FloodBrain, which automates flood impact reporting to expedite response times~\citep{colverd2023floodbrain}, and Llama2-based models designed for emergency classification and issuing public instructions during emergencies when 911 systems become overwhelmed~\citep{otal2024llm}. Further, fine-tuned LLMs have improved disaster-related tweet classification for event identification and aid distribution~\citep{yin2024crisissense}, while other efforts have optimized LLM performance through prompt engineering techniques integrating textual and satellite data~\citep{chen2024optimizing}. DisasterResponseGPT generates personalized disaster preparedness plans, enhancing response efficiency~\citep{goecks2023disasterresponsegpt}.

Several climate-specific LLMs have also emerged. ClimateGPT-2 fine-tunes GPT-2 with climate-specific data to support claim generation and fact-checking tasks~\citep{vaghefi2022deep}. The Arabic Mini-ClimateGPT focuses specifically on conversational Arabic climate instruction tuning using the Clima500-Instruct dataset~\citep{mullappilly2023arabic}, while ChatClimate~\citep{vaghefi2023chatclimate} grounds climate-change-related responses in IPCC AR6 reports, enhancing accuracy and timeliness in climate conversations. Similarly, ClimateGPT synthesizes interdisciplinary climate knowledge through domain-specific instruction tuning and extensive climate-related resource training~\citep{thulke2024climategpt}. Despite their advantages, retraining and fine-tuning these LLMs are computationally intensive, leading to significant environmental impacts due to high carbon emissions. \\ 

\noindent \textbf{Small Language Models (SLMs) alignment.}
Recent studies have focused on reasoning-based alignment methods to enhance small models using larger counterparts' reasoning abilities. Reasoning alignment can significantly improve the performance of smaller models by transferring advanced reasoning from larger models~\citep{ranaldi2024aligning}. Chain-of-Thought reasoning techniques have demonstrated substantial improvements in generalization capabilities~\citep{wu2025cot}, while other research has successfully transferred intermediate reasoning steps to enhance decision-making processes~\citep{ho2022large}. Knowledge Distillation (KD), a popular technique, reduces model complexity and latency while retaining accuracy~\citep{hinton2015distilling, sanh2019distilbert}. The Generalized Knowledge Distillation (GKD) framework extends traditional KD by incorporating structured knowledge to improve training outcomes~\citep{agarwal2024policy}.

Unlike conventional KD, our approach explicitly transfers intermediate reasoning structures rather than solely matching model outputs. Although SLM alignment has been explored extensively within general NLP contexts, its application specifically to extreme weather analysis remains understudied. To address this gap, we propose Extreme Weather Reasoning-Aware Alignment (EWRA), a method that aligns smaller models with larger LLMs by training them on reasoning paths generated by the latter. EWRA enables SLMs to decompose complex problems into structured reasoning steps and produce logical explanations, making them highly specialized for extreme weather analysis tasks. This approach offers computational efficiency advantages over traditional methods, substantially reducing the environmental footprint associated with large-scale model training. Additionally, we deliberately refrain from employing retrieval-augmented generation (RAG)-based systems~\citep{gao2023retrieval,vaghefi2023chatclimate}, as fine-tuning inherently incorporates real-time, domain-specific information essential for handling nuanced queries related to extreme weather events thus limiting hallucination in response more effectively than retrieval-based approaches.

%%%%%%%%%%%%%%%%%%%%%%%%%%%%%%%%%%%%%%%
% METHODOLOGY
%%%%%%%%%%%%%%%%%%%%%%%%%%%%%%%%%%%%%%%
\section{Methodology}
\label{sec:methods}

In this section, we present the ClimaEmpact framework, composed of the ExtremeWeatherNews dataset (section~\ref{sec:methods-dataset}), forming the foundation of our analysis; ExtremeAlign (section~\ref{sec:methods-alignment}), a task-specific alignment dataset enabling SLMs to navigate complex reasoning tasks; and EWRA (section~\ref{sec:methods-ewra}), a reasoning-aware fine-tuning method transferring reasoning capabilities from LLMs to SLMs. 
EWRA improves SLMs' ability to handle ambiguities and complexities inherent in extreme weather information.

%%%%%%%%%%%%%%%%%%%%%%%%%%%%%%%%%%%%%%%
% DATASET
%%%%%%%%%%%%%%%%%%%%%%%%%%%%%%%%%%%%%%%
\subsection{ExtremeWeatherNews Dataset}
\label{sec:methods-dataset}

The ExtremeWeatherNews dataset comprises news articles collected for 60 distinct extreme weather events, and chosen based on ClimaMeter~\citep{faranda2024climameter}\footnote{https://www.climameter.org}. The extreme weather events include heatwaves, cold spells, extreme wind, and extreme precipitation.

To build ExtremeWeatherNews, we employed a web scraping approach, based on Google News RSS feeds and the newspaper3k Python library\footnote{https://newspaper.readthedocs.io/en/latest/}, targeting news articles related to the 60 extreme weather events of interest.
The scraper retrieved RSS feeds within a one-month time window before and after each event date, parsing content to extract relevant details such as article titles, descriptions, and full text, with a focus on English-language news articles. 
To isolate location-specific information, we first extracted sentences from the news articles and then employed the Flair NER tagger\footnote{https://huggingface.co/flair/ner-english} to identify and extract entities of the GPE (Geopolitical Entity) category.
Sentences lacking location tags were considered noisy and subsequently filtered out, resulting in a curated dataset of 127,454 sentences.
This process allowed us to focus on extreme weather-specific news, containing information about vulnerabilities, impacts, and emergency responses across diverse geographical locations. 
Tab.~\ref{tab:data_vis} provides details on the selected events, including event types, dates, and associated locations.

We note that effectively retrieving news articles for a specific extreme weather event requires more than simply searching for a single attribute, such as the event name, since many articles either omit explicit mentions or discuss broader contexts. 
To address this issue, we constructed comprehensive search queries combining the event's name or an appropriate proxy (e.g., ``Typhoon Yagi'' or ``Central Asia heatwave''), the event location obtained from ClimaMeter, and the region of interest for the analysis, along with predefined thematic keywords categorized into: (a) \textit{public}, (b) \textit{economic}, and (c) \textit{weather conditions}. 
Public keywords (e.g., ``relief'', ``evacuation'') emphasize community resilience and emergency responses, economic keywords (e.g., ``economic cost'', ``insurance'') target financial impacts and recovery, and weather conditions keywords (e.g., ``high temperature'', ``rain'') capture specific meteorological factors. 
A comprehensive list of adopted keywords is provided in Tab.~\ref{tab:keywords}.

An example of the search query is: ``\textit{Typhoon Yagi Vietnam resilience weather}'', where we include the general keyword ``\textit{weather}'' to improve relevance and context. 
Without this keyword, terms like ``impact'', ``resilience'', or ``insurance'' could yield results unrelated to weather, such as economic policies or general disaster recovery efforts.
We also note that the time window adopted for our web scraping -- i.e., $\pm$ 1 month with respect to a given extreme weather event -- further anchors the search to relevant content specific to the event of interest.

%%%%%%%%%%%%%%%%%%%%%%%%%%%%%%%%%%%%%%%
% ALIGNMENT DATA
%%%%%%%%%%%%%%%%%%%%%%%%%%%%%%%%%%%%%%%
\subsection{Alignment Data}
\label{sec:methods-alignment}

The dataset introduced in section~\ref{sec:methods-dataset} provides event-specific textual data, serving as a foundation for developing \textit{ExtremeAlign}, a high-quality, task-specific alignment dataset leveraging LLMs. 
Training small language models to effectively handle diverse extreme weather analysis tasks necessitates extensive, accurately annotated data. However, manual annotation of large-scale datasets focused on extreme weather events is both labor-intensive and costly, making it impractical. 
To overcome this challenge, we utilize \textit{Qwen2.5-32B-Instruct}~\citep{qwen2.5}, an expert-level model in extreme weather reasoning, to automatically generate high-quality alignment data through a one-shot prompting strategy, collectively denoted as \( \mathcal{D}_{\text{one-shot}} \).
This approach incorporates comprehensive definitions of task-specific categories and clear guidelines within each prompt, enhancing data consistency, facilitating human alignment, and reducing subjective biases during dataset construction.

To effectively analyze extreme events, we design prompts for three interconnected tasks: \textit{vulnerability/impact/emergency assessment}, \textit{topic/subtopic labeling and keyword extraction}, and \textit{emotion analysis}. 
The first assesses event severity and urgency, the second categorizes thematic content, and the third evaluates public sentiment and emotional responses. 
Together, these tasks provide a comprehensive framework capturing the vulnerability and impact, societal reactions, and broader implications of extreme weather.

Consequently, we formulate the overall dataset generation task as:
\[
\mathcal{D}_{\text{gen}} \leftarrow LLM_p(T, \mathcal{D}_{\text{one-shot}})
\]
where \( \mathcal{D}_{\text{gen}} \) represents the generated dataset, \( p \) indicates the prompt used during inference, and \( T \) refers to the three tasks.

Prior to prompt design, we considered several key aspects to ensure systematic categorization and consistency. 
Clear category definitions, including explicit inclusion and exclusion criteria, were established to avoid ambiguity. 
We also integrated structured reasoning into our prompts, requiring the model to articulate step-by-step thinking before assigning probabilities. 
To enforce interpretability and consistency, we constrained probability scores to sum exactly to one and adopted a strict output format with distinct \texttt{<think>} and \texttt{<output>} sections. A strict format standardizes output structure, which is essential for post-processing and evaluation.
These considerations facilitated structured decision-making, improved response accuracy, and enhanced model reliability in real-world extreme weather analysis.

We followed a consistent prompting strategy across all three tasks:\\
\begin{center}
\doublebox{
  \begin{minipage}{0.9\textwidth}
  \textbf{Step 1:} Given a sentence, we prompt the LLM to identify applicable categories based on the information taxonomy in Tab.~\ref{tab:climate_definitions}. \\
  \textbf{Step 2:} The LLM assigns a probability score (between 0 and 1) to each category, reflecting its confidence. \\
  \textbf{Step 3:} Probability scores across all categories must sum exactly to 1.
  \end{minipage}
}
\end{center}
\vspace{0.3cm}
By framing these tasks as ranking problems, we enable the LLM to learn relative importance among categories, thereby reflecting the inherent ambiguity and complexity of extreme event-related information. 
Generating probability scores instead of single labels allows for representations of uncertainty, accommodates overlapping categories, and supports accurate decision-making in critical applications such as disaster response, risk analysis, and resource allocation. 
Detailed prompts for each task are provided in section~\ref{sec:prompt_details}.

The final dataset consists of 30,000 high-quality annotations, evenly divided across the three primary tasks, ensuring balanced supervision for multi-task alignment.

%%%%%%%%%%%%%%%%%%%%%%%%%%%%%%%%%%%%%%%
% EWRA
%%%%%%%%%%%%%%%%%%%%%%%%%%%%%%%%%%%%%%%
\subsection{Extreme Weather Reasoning-Aware Alignment (EWRA)}
\label{sec:methods-ewra}
\begin{figure}[htbp!]
    \centering
    \includegraphics[width=\linewidth]{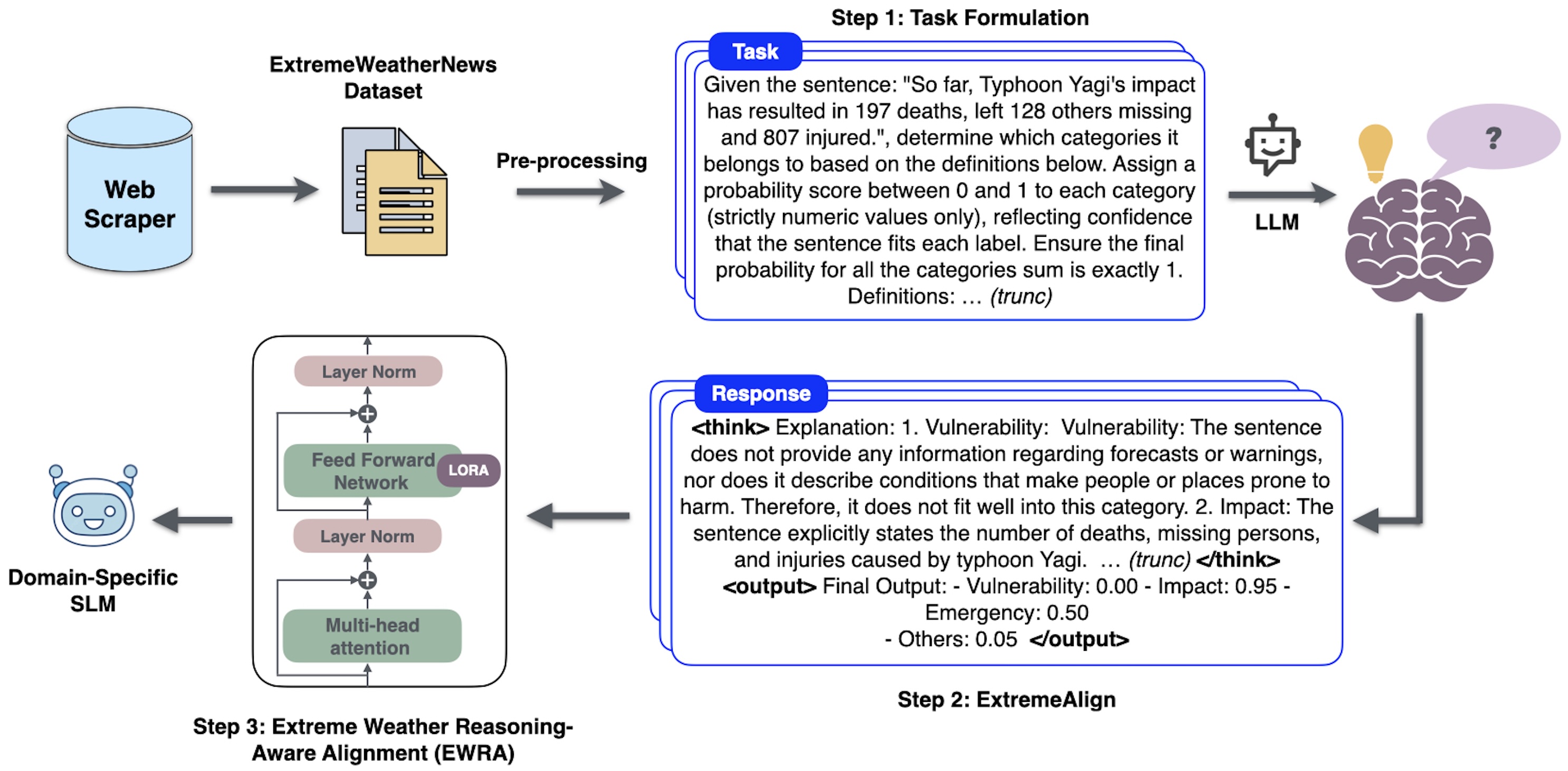}
    \caption{An overview of EWRA. The process begins with web scraping news articles to construct the ExtremeWeatherNews, followed by pre-processing. In Step 1, the task is formulated for an LLM. Step 2 introduces the ExtremeAlign (Alignment Data Construction), which structures reasoning-based outputs. Finally, in Step 3, a domain-specific Small Language Model (SLM) is trained using EWRA to improve reasoning alignment for extreme weather analysis.}
    \label{fig:method}
\end{figure}

To enhance the performance of SLMs, we propose EWRA, an approach that transcends traditional supervised fine-tuning (SFT), which often depends on human-annotated datasets. Instead, EWRA introduces an advanced reasoning-based learning framework, a specialized variant of SFT, paired with a  two-stage curriculum based reasoning transfer mechanism from LLMs. 
Fig.~\ref{fig:method} provides an overview of this process.
%
% \textcolor{red}{GM: Keane, Deeksha, can we significantly improve this figure. It would be good to use stuff from Typhoon Yagi, given that it is the example you use in the introduction.

In this strategy, we train the SLM using a synthetic alignment dataset \( \mathcal{D} = \{(p, q, r_p)\} \), generated by the LLM as described in section \ref{sec:methods-alignment}. 
Here, \( p \) denotes a prompt, \( q \) refers to an input query, and \( r_p \) is the corresponding output that incorporates reasoning, including both intermediate reasoning steps and the final task-specific output. 
The output \( r_p \) is composed of two parts: the reasoning steps \( r_{\text{reasoning}} = [t_1, t_2, \dots, t_k] \), where \( k \) is the total number of reasoning tokens, and the final output \( r_{\text{final}} \), which is the ultimate answer or prediction. The reasoning process is generated step-by-step as \( r_p = [r_{\text{reasoning}}, r_{\text{final}}] \), where \( r_{\text{reasoning}} \) contains intermediate reasoning and \( r_{\text{final}} \) is the final output of the task.

At each time step \( j \), the token \( t_j \) is sampled from the generation distribution $\pi_\phi(\cdot \mid s_j)$, where $\pi_\phi$ represents the model’s learned distribution over possible tokens, and \( s_j \) represents the model's state. 
The training objective is to maximize the likelihood of the reasoning sequence conditioned on the prompt.
We consider two distinct prompting regimes: \begin{inparaenum} \item Explicit prompt incorporates detailed task definitions, inclusion/exclusion criteria, and step-by-step reasoning instructions. These promote structured alignment and precise instruction following. 
\item Implicit prompt, in contrast, excludes such definitions. This setting encourages reliance on the LLM's internalized domain knowledge and captures more flexible, context-dependent reasoning patterns.
\end{inparaenum}

The respective training objectives are defined as:
\begin{equation}
L_{\text{Explicit}}(\phi) = -\mathbb{E}_{(p, q, r_p) \sim \mathcal{D}_{\text{explicit}}} \left[ \sum_{j=1}^{k} \log \pi_\phi(t_j \mid s_j, p) \right],
\end{equation}
\begin{equation}
L_{\text{Implicit}}(\phi) = -\mathbb{E}_{(p, q, r_p) \sim \mathcal{D}_{\text{implicit}}} \left[ \sum_{j=1}^{k} \log \pi_\phi(t_j \mid s_j, p) \right].
\end{equation}

Our training strategy for \textit{EWRA} follows a two-stage curriculum. We begin by training SLMs on data generated with implicit prompts. This stage promotes generalizable, context-aware reasoning by encouraging the model to draw upon latent knowledge from the LLM. We then fine-tune the model using explicitly guided examples that include structured definitions and intermediate reasoning steps. This phase sharpens the model's ability to follow prompts and improves task-specific performance.

This curriculum style approach enables robust transfer of reasoning capabilities from LLMs to SLMs, blending the flexibility of implicit reasoning with the precision of explicit guidance. By leveraging both types of prompts, EWRA supports nuanced decision-making and alignment in high-stakes applications such as extreme weather analysis.

Overall, EWRA leverages reasoning-based alignment to enhance the reasoning capabilities of SLMs for extreme weather analysis. By utilizing ExtremeAlign, a structured alignment dataset generated using LLMs, we transfer domain-specific reasoning properties to SLMs. In the following section, we present our experimental results, demonstrating the effectiveness of EWRA in improving reasoning performance for extreme weather event analysis tasks. 

%%%%%%%%%%%%%%%%%%%%%%%%%%%%%%%%%%%%%%%
% RESULTS
%%%%%%%%%%%%%%%%%%%%%%%%%%%%%%%%%%%%%%%
\section{Experiments}
\label{sec:results}

%%%%%%%%%%%%%%%%%%%%%%%%%%%%%%%%%%%%%%%
% SETUP
%%%%%%%%%%%%%%%%%%%%%%%%%%%%%%%%%%%%%%%
\subsection{Model Setup}
\label{sec:results-setup}
We use Unsloth\footnote{http://github.com/unslothai/unsloth} for training our models and use Unsloth's pre-selected defaults for fine-tuning such as we set the LoRA rank to 16 and fine-tune all modules. 
The LoRA scaling factor is set to 16, with a dropout rate of 0 and no bias applied. 
Gradient checkpointing is enabled using ``unsloth'' for memory efficiency. 
The random seed is fixed at 3407 for reproducibility. Rank-stabilized LoRA is disabled, and loftq config is set to None. 
All the models are trained for one epoch with a learning rate of 2e-4 and a weight decay of 0.01. 
We employ a linear learning rate scheduler with 5 warmup steps. 
The effective batch size is 64, achieved by setting the per-device batch size to 16 and applying gradient accumulation over 4 steps.
Training is optimized using AdamW 8-bit, with mixed precision enabled via FP16 or BF16, depending on hardware support. 
Max sequence length is set to 2048 with 4bit quantization to reduce memory usage. 
Learning rate scheduler type is linear. 
The experiments were conducted on a workstation equipped with four NVIDIA RTX 3090/4090 GPUs, each with 24GB of VRAM. 
For alignment data generation, Qwen2.5-32B-Instruct processes each sample in 1 minute, utilizing 21GB of VRAM. 
Task-specific models were trained with a batch size of 16, a maximum sequence length of 256, a learning rate of 2e-5, and a dropout rate of 0.3. 
For computing BERTScore, we use the default model employed \textit{viz.} roberta-large.
We adopt Query-Key (QK) fine-tuning, where only the query and key matrices in the attention layers are updated \citep{goyal2025context}. Our goal is to align SLMs to reason better about extreme weather events without overwriting their general language understanding. Since the value matrices and MLP layers are key to storing factual knowledge, freezing them preserves the LLM’s pretraining, while still allowing adaptation of attention dynamics.
For our training protocol, we run each of the explicit and implicit alignment datasets for 2 epochs. For the EWRA strategy, we adopt a two-stage curriculum: the model is first trained for 1 epoch on the implicit data, followed by 1 epoch on the explicit data.

%%%%%%%%%%%%%%%%%%%%%%%%%%%%%%%%%%%%%%%
% EVALUATION
%%%%%%%%%%%%%%%%%%%%%%%%%%%%%%%%%%%%%%%
\subsection{Evaluation}
\label{sec:results-evaluation}

Given the subjective nature of extreme weather analysis tasks, we use ranking-based evaluation instead of absolute accuracy to better capture model performance for our task, where multiple correct outputs exist. 
Specifically, we employ Spearman Rank Correlation~\citep{zar2005spearman} to assess alignment between model predictions and human annotations, ensuring relevance and expert-aligned ordering.
\paragraph{Spearman Rank Correlation}
The Spearman rank correlation (SRC) coefficient ($\rho$) measures the strength and direction of the monotonic relationship between predicted and ground-truth rankings. 
For two ranked sets $X$ and $Y$ with $n$ observations, $\rho$ is computed as:
\[
\rho = 1 - \frac{6 \sum_{i=1}^{n} d_i^2}{n(n^2 - 1)}
\]
where $d_i$ represents the rank difference between the $i$-th observation in $X$ and $Y$. 
Spearman’s $\rho$ ranges from -1 (perfect negative correlation) to +1 (perfect positive correlation), with 0 indicating no monotonic association.

Similarly, to evaluate the generated explanations, we use the Jaccard Index, which quantifies the overlap between two sets of words or tokens representing the explanations. 
It evaluates how similar the content of two explanations is by comparing their shared elements. 
It is calculated as the ratio of the number of common words (or tokens) between two explanations to the total number of unique words across both explanations. 
We also report BERTScore\citep{zhang2019bertscore}, which is a token-level similarity metric that evaluates the quality of generated text by computing contextualized embeddings using a pre-trained transformer model. 
Instead of relying on exact word matching, BERTScore measures semantic similarity between candidate and reference texts by computing cosine similarity between their embeddings. 
The range for both metrics is from 0 to 1.

%%%%%%%%%%%%%%%%%%%%%%%%%%%%%%%%%%%%%%%
% MODELS
%%%%%%%%%%%%%%%%%%%%%%%%%%%%%%%%%%%%%%%
\subsection{Model Description}  
\label{sec:results-models}

We compare the performance of different models on the \textit{ExtremeWeatherNews} dataset across three key tasks:  
\begin{inparaenum}[(1)]  
    \item Vulnerability/Impact/Emergency Statement Assessment,  
    \item Topic/Subtopic Labeling and Keyword Extraction,
    \item Emotion Analysis.
\end{inparaenum}  
These tasks are evaluated using the alignment data described in section~\ref{sec:methods-alignment}.

Below, we describe the models used in our experiments.  

\begin{enumerate}
\item{Task-Specific Models:}  
For the first task, we use a RoBERTa-based model~\citep{liu2019roberta} and perform multi-class classification. As this task has not been previously explored, we utilize the model as a baseline for our experiments. For emotion analysis, we leverage the proposed model by \citep{arora2024evaluating} for our experiments. 
We also record the probabilities over all the classes. 
Similarly, for the third task, we also employ RoBERTa, since this task has not been previously formulated as a ranking problem, while for keyword extraction, we utilize concept parsing~\citep{ong2025esgsenticnet,cambria2022sentic} which utilizes part-of-speech (POS) tagging and dependency parsing to identify syntactic relationships. 
This model is a fine-tuned transformer-based classifier optimized individually for each task.
It does not leverage chain-of-thought explanations but is trained specifically for classification using supervised learning.  

\item{Qwen2.5 Instruct Models (1.5B, 3B):}  
Qwen2.5 \citep{qwen2.5} is a family of large language models (LLMs) developed by Alibaba’s Qwen team. 
We experiment with two variants: 
Qwen2.5-1.5B-Instruct, and Qwen2.5-3B-Instruct. 
These models differ in capacity, with larger variants expected to capture more complex reasoning patterns and provide more reliable explanations.  
\begin{enumerate}
    \item {Zero-shot setting}: We established a baseline by performing zero-shot inference on the ExtremeAlign dataset using the Qwen2.5-1.5B-Instruct model without any fine-tuning. This provides a reference point to assess the effectiveness of our alignment strategies.
    \item {Direct-SFT}: A standard supervised fine-tuning (SFT) approach where the model is trained only on the final task-specific outputs (category labels). While category definitions are included in the instruction, the training data omits any reasoning traces or intermediate steps.
    
    \item {ReasonImplicit-SFT}: In this setting, the model is trained on both reasoning traces and final category labels. However, task-specific definitions are excluded from the instructions. This design encourages the model to rely on its internal parametric knowledge and domain understanding during reasoning.
    
    \item {ReasonExplicit-SFT}: The model is trained using both step-by-step reasoning traces and final outputs, with the instructions explicitly prompting for structured reasoning. Additionally, detailed task definitions are included in the prompt to guide the model's predictions.
    
    \item {EWRA (Extreme Weather Reasoning-Aware Alignment)}: Our proposed alignment strategy described in section~\ref{sec:methods-ewra}. It employs a two-stage curriculum. First, the model is trained on implicitly prompted data (without definitions) to encourage generalizable and context-aware reasoning. This is followed by fine-tuning on explicitly guided data (with structured definitions and reasoning), promoting task adherence and instruction following.
    
\end{enumerate}

\end{enumerate}

\subsection{Data Splits and Gold TestSet}
The ExtremeAlign dataset is partitioned into training, validation, and test splits using a 70\%, 15\%, and 15\% ratio, respectively. 
To evaluate model performance under realistic human-like reasoning, we additionally construct a gold-standard test set comprising 600 samples manually annotated by human experts, 200 for each core task: vulnerability/impact/emergency assessment, topic/subtopic/keyword labeling, and emotion analysis. 
The annotation process followed a detailed set of guidelines, as described in section~\ref{sec:annot_guide}. 
These annotations reflect both explicit instruction adherence and implicit domain knowledge, capturing the nuanced decision-making typically employed by humans in extreme weather analysis.

%%%%%%%%%%%%%%%%%%%%%%%%%%%%%%%%%%%%%%%
% QUANTITATIVE RESULTS
%%%%%%%%%%%%%%%%%%%%%%%%%%%%%%%%%%%%%%%
\subsection{Results \& Discussion}
\label{sec:results-quantitative}
Tabs.~\ref{tab:results_1}, \ref{tab:results_2} and~\ref{tab:results_3} report the performance of various models on the ExtremeAlign dataset across three tasks. 
Evaluation is conducted using Spearman's rank correlation coefficient ($\rho$) to assess ranking accuracy, while explanation quality is measured using the Jaccard Index and BERTScore by comparing model outputs with the gold set. 
\begin{table*}[ht!]
\centering
% \small
\caption{Performance comparison across Vulnerability/Impact/Emergency Assessment task using different fine-tuning strategies on Qwen2.5-1.5B and Qwen2.5-3B models. Metrics include Spearman Rank Correlation (SRC ($\rho$)), Jaccard Index, and BERTScore.}
\renewcommand{\arraystretch}{1.0}
\begin{adjustbox}{max width=1.0\textwidth}
\begin{tabular}{|>{\centering\arraybackslash}m{3.3cm}|>{\centering\arraybackslash}m{4.4cm}|>{\centering\arraybackslash}m{2.6cm}|>{\centering\arraybackslash}m{2.6cm}|>{\centering\arraybackslash}m{2.6cm}|}
\hline
\multicolumn{5}{|c|}{\textbf{Vulnerability/Impact/Emergency Assessment}} \\
\hline
\textbf{Model} & \textbf{Method} & \textbf{SRC ($\rho$)} & \textbf{Jaccard Index} & \textbf{BERTScore} \\
\hline
\multirow{1}{*}{\shortstack{Task-Specific Models}} & -  & 0.1467 & - & - \\
\hline
\multirow{5}{*}{\shortstack{Qwen2.5-1.5B- \\Instruct}} & Zero-shot & 0.3638 & 0.1277 & 0.7133  \\
 & Direct-SFT  & {0.7290} & - & - \\
& ReasonImplicit-SFT  &  0.4200 & 0.1861 & 0.7290\\

 & ReasonExplicit-SFT  & \textbf{0.7960 }& \textbf{0.2585} & 0.8897 \\
 & EWRA  & 0.7140 & { 0.2573} & \textbf{0.9028 }\\
\hline
 \multirow{5}{*}{\shortstack{Qwen2.5-3B- \\Instruct}} & Zero-shot & 0.5400 & 0.1226 &  0.3744 \\
  & Direct-SFT  & 0.7910 & - & -  \\
 & ReasonImplicit-SFT  &  0.4860 & 0.1937 &  0.6993 \\
& ReasonExplicit-SFT  & 0.7820 & {0.3294} &  \textbf{0.9163}  \\
& EWRA  & \textbf{0.8230} & \textbf{0.3340} & { 0.9161} \\
\hline
\end{tabular}
\end{adjustbox}
\label{tab:results_1}
\end{table*}

\begin{table*}[ht!]
\centering
% \small
\caption{Performance comparison across Topic/Subtopic Labeling and Keyword Extraction task, using different fine-tuning strategies on Qwen2.5-1.5B and Qwen2.5-3B models. Metrics include Spearman Rank Correlation (SRC ($\rho$)), Jaccard Index, and BERTScore.}
\renewcommand{\arraystretch}{1.0}
\begin{adjustbox}{max width=1.0\textwidth}
\begin{tabular}{|>{\centering\arraybackslash}m{3.3cm}|>{\centering\arraybackslash}m{4.4cm}|>{\centering\arraybackslash}m{2.6cm}|>{\centering\arraybackslash}m{2.6cm}|>{\centering\arraybackslash}m{2.6cm}|}
\hline
\multicolumn{5}{|c|}{\textbf{Topic/Subtopic Labeling}} \\
\hline
\textbf{Model} & \textbf{Method} & \textbf{SRC ($\rho$)} & \textbf{Jaccard Index} & \textbf{BERTScore} \\
\hline
\multirow{1}{*}{\shortstack{Task-Specific Models}} & - &  0.2100/0.1999 & - & - \\
\hline
\multirow{5}{*}{\shortstack{Qwen2.5-1.5B- \\Instruct}} & Zero-shot & 0.4104/\textbf{0.6702} & 0.0436/0.0436 & 0.6342/0.6342  \\
 & Direct-SFT  & 0.3264/0.5770 & -/- & -/-  \\
& ReasonImplicit-SFT & 0.3040/0.52883 & 0.1670/0.1670 & 0.8741/0.8741  \\

 & ReasonExplicit-SFT  & {0.5157/0.6589} & \textbf{0.1744/0.1720} & \textbf{0.8772/0.8772}  \\
 & EWRA  & \textbf{0.5673}/0.6671& 0.1711/0.1711 & 0.8765/0.8765   \\
\hline
 \multirow{5}{*}{\shortstack{Qwen2.5-3B- \\Instruct}} & Zero-shot & 0.4660/\textbf{0.7727} & 0.1592/0.1592 & 0.8257/0.8257\\
  & Direct-SFT & 0.4179/0.6397 &  -/- & -/-\\
 & ReasonImplicit-SFT & 0.4131/0.6066 & 0.1907/0.1671 & 0.8831/0.8741 \\
& ReasonExplicit-SFT  & \textbf{0.5692}/0.6527 & \textbf{0.2045}/0.1743 & 0.8879/0.8771  \\
& EWRA  & 0.5386/0.7149 & 0.2026/\textbf{0.2026} & \textbf{0.8890/0.8890} \\
\hline
% \hline
\multicolumn{5}{|c|}{\textbf{Keyword Extraction}} \\
\hline
\textbf{Model} & \multicolumn{2}{c|}{\textbf{Method}} & \multicolumn{2}{c|}{\textbf{Jaccard Index}} \\
\hline
\multirow{1}{*}{\shortstack{Task-Specific Models}} & \multicolumn{2}{c|}{-} & \multicolumn{2}{c|}{0.1321} \\
\hline

\multirow{5}{*}{\shortstack{Qwen2.5-1.5B- \\Instruct}} & \multicolumn{2}{c|}{Zero-shot} &  \multicolumn{2}{c|}{0.1257} \\
 & \multicolumn{2}{c|}{Direct-SFT}  & \multicolumn{2}{c|}{0.1984} \\
& \multicolumn{2}{c|}{ReasonImplicit-SFT} &  \multicolumn{2}{c|}{\textbf{0.2813}} \\

 & \multicolumn{2}{c|}{ReasonExplicit-SFT}  & \multicolumn{2}{c|}{{0.2343}} \\
 & \multicolumn{2}{c|}{EWRA } & \multicolumn{2}{c|}{0.2333} \\
\hline
 \multirow{5}{*}{\shortstack{Qwen2.5-3B- \\Instruct}} & \multicolumn{2}{c|}{Zero-shot} &  \multicolumn{2}{c|}{0.2817} \\
  & \multicolumn{2}{c|}{Direct-SFT}  & \multicolumn{2}{c|}{0.2698} \\
 & \multicolumn{2}{c|}{ReasonImplicit-SFT} & \multicolumn{2}{c|}{\textbf{0.3269 }}\\
& \multicolumn{2}{c|}{ReasonExplicit-SFT}  &  \multicolumn{2}{c|}{{0.2343}} \\
& \multicolumn{2}{c|}{EWRA } & \multicolumn{2}{c|}{0.2856} \\
\hline
\end{tabular}
\end{adjustbox}
\label{tab:results_2}
\end{table*}

Overall, EWRA consistently outperforms other training strategies, particularly when applied to the Qwen2.5-3B-Instruct model with higher number of parameters. 
Compared to ReasonExplicit-SFT, EWRA provides clear improvements across most tasks and metrics. 
For instance, in the Vulnerability/Impact/Emergency assessment task on the 3B model, EWRA improves Jaccard Index by 2.7\% and increases SRC score by 4.3\%. 
These gains indicate that reasoning alignment helps models generalize better and make more precise predictions. EWRA and ReasonExplicit-SFT are the top-performing fine-tuning strategies overall. 
While improvements in generated explanation (measured via BERTScore) are less significant as compared to ReasonExplicit-SFT, this is likely due to the fact that these metrics emphasize semantic overlap, which may remain high even when explanation structure or specificity differs. 
As such, BERTScore may under-represent the qualitative differences in reasoning alignment captured by EWRA.
\begin{table*}[ht!]
\centering
% \small
\caption{Performance comparison across Emotion Analysis task using different fine-tuning strategies on Qwen2.5-1.5B and Qwen2.5-3B models. Metrics include Spearman Rank Correlation (SRC ($\rho$)), Jaccard Index, and BERTScore.}
\renewcommand{\arraystretch}{1.0}
\begin{adjustbox}{max width=1.0\textwidth}
\begin{tabular}{|>{\centering\arraybackslash}m{3.3cm}|>{\centering\arraybackslash}m{4.4cm}|>{\centering\arraybackslash}m{2.6cm}|>{\centering\arraybackslash}m{2.6cm}|>{\centering\arraybackslash}m{2.6cm}|}
\hline
\multicolumn{5}{|c|}{\textbf{Emotion Analysis}} \\
\hline
\textbf{Model} & \textbf{Method} & \textbf{SRC ($\rho$)} & \textbf{Jaccard Index} & \textbf{BERTScore} \\
\hline
\multirow{1}{*}{\shortstack{Task-Specific Models}} & - & 0.2876 & - & - \\
\hline
\multirow{5}{*}{\shortstack{Qwen2.5-1.5B- \\Instruct}} & Zero-shot & 0.6071 & 0.1336 & 0.7434  \\
 & Direct-SFT   & \textbf{0.8389} & - & -  \\
& ReasonImplicit-SFT  & 0.7828 & 0.2528 & \textbf{0.8986} \\
 & ReasonExplicit-SFT  & 0.5578 & \textbf{0.2583} & 0.8979 \\
 & EWRA  & 0.8308 & 0.2565 & 0.8976 \\
\hline
 \multirow{5}{*}{\shortstack{Qwen2.5-3B- \\Instruct}} & Zero-shot & 0.6458 & 0.1112 & 0.2666 \\
  & Direct-SFT  & 0.8442 & - &  - \\
 & ReasonImplicit-SFT:  & 0.8380 & 0.2627 & \textbf{0.9011} \\
& ReasonExplicit-SFT  & \textbf{0.8716} & 0.2646 & {0.8989} \\
& EWRA  & {0.8708} & \textbf{0.2664} & 0.8995 \\
\hline
\end{tabular}
\end{adjustbox}
\label{tab:results_3}
\end{table*}

\begin{enumerate}
    \item Zero-shot vs EWRA:  zero-shot models typically underperform compared to EWRA and other fine-tuned variants—due to their lack of task-specific alignment and limited domain grounding. However, we may note that Qwen2.5-1.5B-Instruct in the zero-shot setting achieved a Spearman correlation of 0.4104/0.6702, while the EWRA-aligned model reached only 0.5673/0.6671. A similar trend is observed for Qwen2.5-3B-Instruct, where the zero-shot model attained 0.4660/0.7727 , outperforming the EWRA variant (0.5386/0.7149). While EWRA is effective for broad topic labeling and structured categorization, it falls short in subtopic labeling, which often requires nuanced semantic understanding and sensitivity to fine-grained distinctions. One likely reason is the task design: subtopic evaluation relies on ranking-based accuracy metrics, which are sensitive to the model's ability to produce all relevant labels. However, SFT-aligned models—particularly under EWRA—tend to predict only a subset of categories, reducing performance on ranking-based evaluations. This is further supported by explanation similarity metrics (e.g., BERTScore), where ReasonExplicit and EWRA settings achieve the highest scores, indicating that while reasoning quality improves, the alignment objective may suppress broader coverage needed for exhaustive subtopic labeling.
    \item ReasonImplicit-SFT vs ReasonExplicit-SFT: A comparison between ReasonImplicit-SFT and ReasonExplicit-SFT highlights the importance of including explicit definitions and step-by-step guidance. While gains in Emotion analysis are relatively small (e.g., BERTScore improvement of ~0.15\%), Vulnerability/Impact/Emergency assessment task shows a 70.3\% relative improvement in Jaccard when moving from implicit to explicit reasoning on the 3B model. This suggests that when categories are fine-grained or semantically overlapping, models benefit from being explicitly told how to structure their reasoning. However, we observe that results on keyword extraction reveal that ReasonImplicit-SFT outperforms ReasonExplicit-SFT in several cases, especially on the smaller Qwen2.5-1.5B model. This can be attributed to the flexible nature of the task: since keyword extraction does not rely on fixed category definitions, models may benefit from a more latent understanding of task intent, which ReasonImplicit-SFT encourages. The implicit formulation likely gives the model room to adapt and use contextual cues from the input without being overly constrained by rigid definitions. This suggests that ReasonImplicit-SFT may promote better domain adaptation in tasks like keyword extraction, where the model must infer task structure from context rather than from explicit templates.
    \item ReasonExplicit-SFT vs Direct SFT: Direct-SFT achieves a high Spearman correlation (e.g., 0.8389 for the 3B model) on the emotion analysis task, likely due to the dominance of a few frequent classes such as sadness, and fear. While this indicates some alignment with ground-truth rankings, the absence of reasoning traces suggests that the model may be overfitting to surface-level patterns or task-specific heuristics rather than learning meaningful structure. In contrast, EWRA  and ReasonExplicit-SFT consistently performs well across all metrics, demonstrating its ability to generate outputs that are not only accurate but also coherent and well-aligned with the reasoning behind the task.
\end{enumerate}

\subsection{Effectiveness of EWRA}
To further examine the role of reasoning alignment in model performance, we perform an ablation study by introducing a ReverseEWRA (Tab.~\ref{tab:results_abl_split}) setting. 
ReverseEWRA is a contrastive variant of EWRA in which the curriculum order is reversed: the model is first trained on explicit instructions and reasoning, followed by fine-tuning using the implicit setting. 
This baseline helps isolate the impact of curriculum ordering on model alignment and reasoning capabilities. 
The motivation behind this experiment is to assess whether the performance gains in EWRA stem from the curriculum structure rather than just prolonged fine-tuning.

\begin{table*}[ht!]
\centering
% \small
\caption{Effectiveness of EWRA: Task-wise Ablation using ReverseEWRA}
\renewcommand{\arraystretch}{1.0}
\begin{adjustbox}{max width=1.0\textwidth}
\begin{tabular}{|>{\centering\arraybackslash}m{4.2cm}|>{\centering\arraybackslash}m{3.4cm}|>{\centering\arraybackslash}m{2.6cm}|>{\centering\arraybackslash}m{2.6cm}|>{\centering\arraybackslash}m{2.6cm}|}
\hline
\multicolumn{5}{|c|}{\textbf{Vulnerability/Impact/Emergency Assessment}} \\
\hline
\textbf{Model} & \textbf{Method} & \textbf{SRC ($\rho$)} & \textbf{Jaccard Index} & \textbf{BERTScore} \\
\hline
Qwen2.5-1.5B-Instruct & ReverseEWRA & 0.7080 & 0.2555 & 0.9025 \\
Qwen2.5-3B-Instruct   & ReverseEWRA & 0.7070 & 0.3117 & 0.9124 \\
\hline
% \multicolumn{5}{c}{} \\
\hline
\multicolumn{5}{|c|}{\textbf{Topic/Subtopic Labeling}} \\
\hline
\textbf{Model} & \textbf{Method} & \textbf{SRC ($\rho$)} & \textbf{Jaccard Index} & \textbf{BERTScore} \\
\hline
Qwen2.5-1.5B-Instruct & ReverseEWRA & 0.5182/0.4949 & 0.1595/0.1595 & 0.8729/0.8728 \\
Qwen2.5-3B-Instruct   & ReverseEWRA & 0.5209/0.6747 & 0.1931/0.1931 & 0.8859/0.8859 \\
\hline
\multicolumn{5}{|c|}{\textbf{Keyword Extraction}} \\
\hline
\textbf{Model} & \multicolumn{2}{c|}{\textbf{Method}} & \multicolumn{2}{c|}{\textbf{Jaccard Index}} \\
\hline
Qwen2.5-1.5B-Instruct & \multicolumn{2}{c|}{ReverseEWRA} & \multicolumn{2}{c|}{0.2411}  \\
Qwen2.5-3B-Instruct   & \multicolumn{2}{c|}{ReverseEWRA} & \multicolumn{2}{c|}{0.2939}  \\
% \multicolumn{5}{c}{} \\
\hline
\hline
\multicolumn{5}{|c|}{\textbf{Emotion Analysis}} \\
\hline
\textbf{Model} & \textbf{Method} & \textbf{SRC ($\rho$)} & \textbf{Jaccard Index} & \textbf{BERTScore} \\
\hline
Qwen2.5-1.5B-Instruct & ReverseEWRA & 0.8650 & 0.2512 & 0.8974 \\
Qwen2.5-3B-Instruct   & ReverseEWRA & 0.8543 & 0.2667 & 0.9005 \\
\hline
\end{tabular}
\end{adjustbox}
\label{tab:results_abl_split}
\end{table*}

Across all tasks, particularly those involving multi-label ranking, such as Vulnerability/Impact/Emergency assessment and Topic/Subtopic labeling, we observe notable performance degradation in the ReverseEWRA setting. 
For example, on the Qwen2.5-3B model, the SRC score for Vulnerability/Impact/Emergency assessment drops from 0.8230 in EWRA to 0.7070 in ReverseEWRA. ReverseEWRA leads to a consistent drop in SRC for Topic/Subtopic labeling across both models, with a relative decline of approximately 8.7\% (from 0.5673 to 0.5182) and 25.8\% (from 0.6671 to 0.4949) for Qwen2.5-1.5B, and 3.3\% (from 0.5386 to 0.5209) and 5.6\% (from 0.7149 to 0.6747) for Qwen2.5-3B.

Interestingly, the Qwen2.5-1.5B model achieves its best performance in Emotion Analysis under the ReverseEWRA setting, with an SRC score of 0.8650—outperforming all other fine-tuning strategies for that task. 
This suggests that in some domains (like emotion inference), starting with explicit reasoning may provide a better inductive bias for low-capacity models, even if the overall curriculum is less optimal across tasks.
%

%%%%%%%%%%%%%%%%%%%%%%%%%%%%%%%%%%%%%%%%%%%%%%%
% CONCLUSIONS
%%%%%%%%%%%%%%%%%%%%%%%%%%%%%%%%%%%%%%%%%%%%%%%
\section{Conclusion}
\label{sec:conclusions}

In this study, we presented Extreme Weather Reasoning-Aware Alignment (EWRA), a novel fine-tuning framework that transfers the structured reasoning capabilities of large language models (LLMs) to small language models (SLMs) for the domain of extreme weather event analysis. 
By integrating LLM-generated reasoning paths into a two-stage training strategy, EWRA enables SLMs to develop both context-aware and task-specific reasoning skills. 
We introduced two new datasets: ExtremeWeatherNews, a rich collection of extreme weather event news data, and ExtremeAlign, a reasoning-augmented alignment set tailored to three key analysis tasks: vulnerability/impact/emergency assessment, topic/subtopic/keyword labeling, and emotion analysis. 
Our experiments across these tasks demonstrate that EWRA substantially improves the alignment of SLMs with domain-specific reasoning, outperforming both standard supervised fine-tuning and other reasoning-based approaches, particularly on models such as Qwen2.5-3B-Instruct with higher number of parameters as compared to Qwen2.5-1.5B-Instruct model. 
In future work, we aim to extend EWRA to broader generative tasks such as temporal summarization and scenario simulation, while also investigating cross-task generalization by developing modular, instruction-tuned variants of ExtremeAlign.

%TC:endignore
% \subsection{Code availability}
% All the relevant codes for this study are provided in a public repository \url{https://github.com/MathEXLab/.....}.

\backmatter

%%%%%%%%%%%%%%%%%%%
% EXTENDED DATA
%%%%%%%%%%%%%%%%%%%
\newpage
% \section*{Extended Data}\label{sec:extended_data}
%
% \renewcommand{\figurename}{Extended Data Fig.}
% \renewcommand{\thefigure}{\arabic{figure}}
% \setcounter{figure}{0}

% \renewcommand{\tablename}{Extended Data Tab.}
% \renewcommand{\thetable}{\arabic{table}}
% \setcounter{table}{0}

% \bmhead{Acknowledgments}
% This work is supported by ...

%% BIBLIOGRAPHY
\bibliography{sn-bibliography}

\renewcommand{\figurename}{Fig.}
\renewcommand{\thefigure}{S\arabic{figure}}
\setcounter{figure}{0}

\renewcommand{\tablename}{Tab.}
\renewcommand{\thetable}{S\arabic{table}}
\setcounter{table}{0}

\newpage 

\appendix
% Redefine subsection numbering for Supplementary Information
\renewcommand{\thesubsection}{S.\arabic{subsection}}

% \section*{Supplementary Information}
% \addcontentsline{toc}{section}{Supplementary Information}

\setcounter{table}{0}
\setcounter{figure}{0}

\section{ExtremeWeatherNews Dataset}
\label{sec:dataset}
% \textcolor{red}{GM: Can we add some text here, describing this appendix and its content? Should we also add more info on the ExtremeWeatherNews, as well as on the Align dataset?}

This appendix section describes the ExtremeWeatherNews dataset and the steps used for data collection. The dataset comprises a curated corpus of news articles and reports centered around major extreme weather incidents. Events were selected using Climameter’s event lists \citep{faranda2024climameter}, ensuring broad geographic and thematic coverage. 
To streamline data scraping, we implemented a multithreaded script capable of handling a high volume of concurrent requests. For targeted retrieval of relevant content, we developed a carefully curated set of keywords (see Tab.~\ref{tab:keywords}). These keywords played a critical role in capturing a comprehensive and diverse range of articles. Tab.~\ref{tab:data_vis} lists the selected events used for news collection and additional details on the selected events, including event types, dates, and associated locations.
To ensure geographical relevance, we applied a filtering process based on location mentions. Specifically, we used administrative codes from the GeoNames\footnote{https://www.geonames.org/} database—namely admin1code (primary divisions such as states or provinces) and admin2code (secondary divisions like cities or districts). We retained only those articles whose mentioned locations matched the geographical scope of the corresponding events. Articles were further filtered to exclude sentences shorter than 30 characters or longer than 200 characters.
Finally, to reduce noise and eliminate content unrelated to extreme weather, human annotators manually reviewed the data. They removed irrelevant samples, particularly those referring to unrelated domains. Further details about the annotation process are provided in Section~\ref{sec:annot_guide}.
\begin{table}[ht!]
    \centering
    \small
    \caption{Keywords for Extreme Weather Event Search} \label{tab:keywords}
    \renewcommand{\arraystretch}{1.2}
\begin{adjustbox}{max width=1.0\textwidth}
    \begin{tabular}{|p{6cm}|p{6cm}|p{6cm}|}
        \toprule
        \textbf{Public Keywords} & \textbf{Economic Keywords} & \textbf{Weather Conditions Keywords} \\
        \midrule
        public, community, people, infrastructure, society, impact, disruption, affected, resilience, support, relief, aid, assistance, emergency, response, preparedness, adaptation, mitigation, awareness, engagement, cooperation, solidarity, social, health, welfare, equity, inclusion, vulnerability, risk, protection, shelter, evacuation, relocation, caution, damage, evacuation, injury, help, sympathy & damage, loss, economic, cost, impact, financial, property, disaster, recovery, reconstruction, insurance, business, investment, job\_loss, economic\_growth, market\_disruption, supply\_chain, infrastructure, resilience, recovery\_funds, economic\_development, employment, gdp\_impact, financial\_aid, bailout, debt, bankruptcy, taxation & dry, snow, high temperature, wind, thunderstorms, rain, winter, cold, summer, hot, lost moisture, pressure, water vapor, sea level pressure, precipitation \\
        \bottomrule
    \end{tabular}
    \end{adjustbox}
\end{table}

\begin{table}[t!]
\centering
% \small
\caption{Summary of Events with Types, Locations, and Dates.} %\textcolor{red}{GM: why does this have only 5 events, and not all 60?}}
\label{tab:data_vis}
    \renewcommand{\arraystretch}{1.0}
\begin{adjustbox}{max width=0.72\textwidth}
\begin{tabular}{|l|l|l|l|}
\hline
\textbf{Event Name}    & \textbf{Event Type} & \textbf{Country}       & \textbf{Event Date} \\ \hline
Romania Floods & floods & Romania & 31/08/2024 \\
\hline
Poland Floods & floods & Poland & 18/08/2024 \\
\hline
USA Winter Storm & cold & USA & 14/01/2024 \\
\hline
Scandinavian Cold Spell & cold & Norway & 08/01/2024 \\
\hline
Ciro Snowstorm & cold & Germany & 02/12/2023 \\
\hline
North American Winter Storm & cold & Canada & 27/12/2022 \\
\hline
Hurricane Helene & wind & USA & 27/09/2024 \\
\hline
Typhoon Yagi & wind & Vietnam & 08/09/2024 \\
\hline
Storm Ingunn & wind & Norway & 01/02/2024 \\
\hline
Cyclone Belal & wind & France & 15/01/2024 \\
\hline
Cyclone Jasper & wind & Australia & 18/12/2023 \\
\hline
Storm Ciaran & wind & France & 03/11/2023 \\
\hline
Cyclone Tej & wind & Yemen & 23/10/2023 \\
\hline
Storms Babet and Aline & wind & Scotland & 20/10/2023 \\
\hline
Storm Poly & wind & Denmark & 05/07/2023 \\
\hline
Hurricane Ian Landfall & wind & USA & 28/09/2022 \\
\hline
April 2020 USA Tornado Outbreak & wind & USA & 12/04/2020 \\
\hline
Hurricane Irma Caribbean Landfall & wind & Caribbean & 07/09/2017 \\
\hline
European Heatwave & heatwave & Albania & 19/07/2024 \\
\hline
Eastern United States Heatwave & heatwave & USA & 23/06/2024 \\
\hline
Saudi Arabia Heatwave & heatwave & Saudi Arabia & 18/06/2024 \\
\hline
Eastern Mediterranean Heatwave & heatwave & Cyprus & 14/06/2024 \\
\hline
India Heatwave & heatwave & India & 29/05/2024 \\
\hline
Easter Extreme Weather in Europe & heatwave & Italy & 01/04/2024 \\
\hline
Morocco Heatwave & heatwave & Morocco & 15/02/2024 \\
\hline
Central Asia Heatwave & heatwave & Pakistan & 30/11/2023 \\
\hline
Brazil Heatwave & heatwave & Brazil & 19/11/2023 \\
\hline
October Heatwave in Europe & heatwave & Switzerland & 13/10/2023 \\
\hline
September Heatwave in Southern and Central Europe & heatwave & France & 10/09/2023 \\
\hline
Late Summer French Heatwave & heatwave & France & 23/08/2023 \\
\hline
Western USA Heatwave & heatwave & USA & 31/07/2023 \\
\hline
Cerberus Heatwave in Southern Europe & heatwave & Italy & 25/07/2023 \\
\hline
Southeast Asia Heat Peak & heatwave & Thailand & 15/04/2023 \\
\hline
Italy Multiple Floods & rain & Italy & 19/10/2024 \\
\hline
Storm Kirk & rain & France & 09/10/2024 \\
\hline
Storm Boris & rain & Austria & 15/09/2024 \\
\hline
Hurricane Beryl & rain & Jamaica & 03/07/2024 \\
\hline
Genoa Low Summer Floods & rain & France & 24/06/2024 \\
\hline
Bavaria Floods & rain & Germany & 03/06/2024 \\
\hline
Texas Floods & rain & USA & 05/05/2024 \\
\hline
South Brazil Floods & rain & Brazil & 02/05/2024 \\
\hline
China Floods & rain & China & 23/04/2024 \\
\hline
Dubai Floods & rain & UAE & 16/04/2024 \\
\hline
Storm Monica & rain & France & 09/03/2024 \\
\hline
California Floods & rain & USA & 01/02/2024 \\
\hline
San Diego Floods & rain & USA & 01/22/2024\\
\hline
North-West USA and Canada Atmospheric River & rain & USA & 06/12/2023 \\
\hline
France and Italy Floods & rain & France & 21/11/2023 \\
\hline
Hurricane Otis & rain & Mexico & 25/10/2023 \\
\hline
New York Floods & rain & USA & 29/09/2023 \\
\hline
Mediterranean Depression Elias & rain & Greece & 27/09/2023 \\
\hline
Cape Town Floods & rain & South Africa & 25/09/2023 \\
\hline
Cevennes Floods & rain & France & 17/09/2023 \\
\hline
Medicane Daniel & rain & Lybia & 11/09/2023 \\
\hline
Guangdong and Hong Kong Floods & rain & Hong Kong & 08/09/2023 \\
\hline
Mediterranean Depression Daniel & rain & Greece & 05/09/2023 \\
\hline
Mediterranean Depression Rea & rain & Italy & 29/08/2023 \\
\hline
Storm Hans in Scandinavia & rain & Norway & 08/08/2023 \\
\hline
California Atmospheric River & rain & USA & 10/01/2023 \\
\hline
Medicane Ianos & rain & Greece & 18/09/2020 \\
\hline
\end{tabular}
\end{adjustbox}
\end{table}

\newpage

\section{Methodology}
\label{sec:method}

% \textcolor{red}{GM: it looks a little short as an appendix -- nothing else to add?}

We begin by developing a robust taxonomy, establishing a foundational structure that organizes climate concepts into key topics, subtopics, and specific entities. 
This taxonomy covers areas such as vulnerabilities, impacts, and emergency responses, creating a data-driven framework that supports integration with broader climate analysis tools. 
By categorizing data across these domains, the taxonomy enables dynamic, multi-dimensional exploration of insights, ranging from localized effects to wider climate patterns. 
However, empirical approaches in the literature are limited \citep{islam2022knowurenvironment,mishra2021neuralnere}, and these methods do not involve human evaluation to ensure the meaningfulness of constructed climate-related topics. 
Consequently, BERTopic is run on climate news articles, with topic keywords provided to climate data experts, to facilitate the development of the taxonomy. 
The taxonomy (Tab.~\ref{tab:climate_definitions}) organizes information into topics and subtopics, making it easy to explore extreme weather data from broad categories down to specific details.
\begin{table*}[ht!]
% \small
\caption{A taxonomical organization of the analysis of extreme weather. Definitions of Extreme Weather  Categories.}
\label{tab:climate_definitions}
\centering
\renewcommand{\arraystretch}{1.2}
\begin{adjustbox}{max width=1.0\textwidth}
\begin{tabular}{|>{\arraybackslash}p{5.5cm}|>{\centering\arraybackslash}p{11cm}|>{\centering\arraybackslash}p{5cm}|}
\hline
\textbf{Topic} & \textbf{Definition} \\ 
\hline
\textbf{Vulnerabilities (V)} & Alert communities based on their specific vulnerability factors  \\ 
\hline
{- Environmental Vulnerability (V)} & Areas prone to greater damage due to fragile ecosystems, with location mention \\ 
\hline
{- Infrastructure Vulnerability (V)} & Structural deficiencies that increase the risk of damage during extreme weather events, with location mention \\ 
\hline
{- Economic Vulnerability (V)} & The susceptibility of an economy to financial losses due to disasters   \\ 
\hline
\textbf{Impact (I)} & Immediate and long-term effects of extreme weather events \\ 
\hline
{- Deaths (I)} & Fatalities caused by extreme weather events \\ 
\hline
{- Infrastructure Damage (I)} & Physical harm to buildings, roads, and other critical structures \\ 
\hline
{- Economic Damage (I)} & Financial losses incurred \\ 
\hline
{- Homeless (I)} & People displaced due to the destruction \\ 
\hline
\textbf{Emergency Response (E)} & Immediate actions taken by authorities, organizations, and communities  \\ 
\hline
{- Evacuation (E)} & Organized movement of people to safety from threatened areas \\ 
\hline
{- Community Support (E)} & Assistance provided to affected populations by both local organizations and broader networks  \\ 
\hline
{- Emergency Services (E)} & Immediate rescue and medical aid delivered during a crisis\\ 
\hline
{- Communication Strategies (E)} & Plans for conveying crucial information to the public and responders during emergencies \\ 
\hline
\end{tabular}
\end{adjustbox}
\end{table*}

\newpage
\subsection{Prompt Details}
\label{sec:prompt_details}
Fig.~\ref{fig:prompt1} shows one-shot prompt for vulnerability/impact/emergency statement assessment.
Similarly, Fig.~\ref{fig:prompt2} and Fig.~\ref{fig:prompt3} shows one-shot prompt for emotion analysis and topic/subtopic/keyword labeling.
\begin{figure}[ht!]
    \centering
    % \small
    \caption{One-shot prompt for vulnerability/impact/emergency statement assessment}
    % \begin{tcolorbox}[colframe=black, colback=white, boxrule=0.8pt, width=\textwidth]
    \resizebox{1.0\textwidth}{!}{
    \begin{tabular}{|p{22cm}|}
    \hline
    [Task Description:] \\
    Given the sentence: "\texttt{<sentence>}", determine which categories it belongs to based on the definitions below. Assign a probability score between 0 and 1 to each category (strictly numeric values only), reflecting confidence that the sentence fits each label. Ensure the final probability for all the categories sum is exactly 1. 

    [Definitions of Categories] \\
    Vulnerability: Describes conditions that make people or places prone to harm, including: \\
    - Forecasts or warnings for specific locations about hazardous conditions (e.g., storm alerts, flood watches). \\
    - Excludes: General weather forecasts without warnings, past rainfall amounts, or climate trends without immediate hazard warnings. \\
    - Special case: Mentions of inches of rain count only if linked to a forecast predicting danger. \\[5pt]
    Impact: Describes strictly measurable consequences of extreme weather, such as: \\
    - Number of casualties, injuries, financial loss, infrastructure damage, or economic impact. \\
    - Excludes: Mentions of states of emergency, inches of rain, road closures, tree falling, or event cancellations without quantifiable impact or flight delays. \\
    - Special case: Casualties always count as Impact, even if estimates (e.g., 'dozens injured'). \\[5pt]
    Emergency: Describes urgent actions requiring immediate response, including: \\
    - Evacuations, rescues, emergency shelters, or disaster response efforts. \\
    - Excludes: State of emergency declarations without mention of direct emergency actions. \\[5pt]
    Others: \\
    - Sentences about extreme weather  without clear Vulnerability, Impact, or Emergency markers. \\
    - Sentences mentioning multiple events without enough detail to fit a single category. \\
    - States of emergency, road closures, or school closures without measurable damage or emergency actions. \\[5pt]

    [Example:] \\
    Input: Krakow is struggling after heavy rainfall, with city officials offering sandbags to protect homes. \\
    \texttt{<think>} \\
        Explanation: \\
        1. Vulnerability: Krakow is struggling after heavy rainfall, indicating the city's vulnerability to flooding. The mention of heavy rainfall implies a risk, suggesting that Krakow is susceptible to potential harm due to the weather conditions. This fits the Vulnerability category as it shows the city’s increased risk of flooding. \\
        2. Impact: While the sentence mentions city officials offering sandbags to protect homes, there is no specific mention of measurable impacts such as casualties, injuries, financial loss, or infrastructure damage. Therefore, this category is less relevant compared to Vulnerability and Emergency. \\
        3. Emergency: The fact that city officials are offering sandbags to protect homes implies an urgent response to a disaster, categorizing the action as an emergency. The immediate need for protective measures points to an emergency situation. \\
        4. Others: This sentence describes a specific event related to heavy rainfall, but it is clearly categorized into Vulnerability, Impact, and Emergency. Therefore, it doesn't fit into the "Others" category as it clearly involves measurable consequences and urgent actions. \\
    \texttt{</think>} \\
    \texttt{<output>} \\
        Final Output: \\
        - Vulnerability: 0.40 \\
        - Impact: 0.10 \\
        - Emergency: 0.50 \\
        - Others: 0.00 \\
    \texttt{</output>} \\
    \hline
    \end{tabular}
    }
    % \end{tcolorbox}
    \label{fig:prompt1}
\end{figure}
\begin{figure}[ht!]
    \centering
    % \small
    \caption{One-shot prompt for emotion analysis.}
    % \begin{tcolorbox}[colframe=black, colback=white, boxrule=0.8pt, width=\textwidth]
    \resizebox{1.0\textwidth}{!}{
    \begin{tabular}{|p{22cm}|}
    \hline
    {[Task Description:]} \\
    Given the sentence: "\texttt{<sentence>}", determine which emotions it conveys. Assign probability scores ensuring they sum to exactly 1. \\

    {[Emotion Definitions]} \\
    {Sadness:} Expresses sorrow, grief, disappointment, or loss, often involving suffering or destruction. \\
    {Anger:} Indicates frustration, outrage, or dissatisfaction, including expressions of blame or criticism. \\
    {Fear:} Suggests concern, anxiety, or perceived threat, often linked to warnings or potential dangers. \\
    {Joy:} Reflects happiness, relief, celebration, or positive outcomes. \\
    {Optimism:} Shows hope, encouragement, or confidence in a positive future. \\
    {Trust:} Indicates reliability, assurance, or faith in a person, system, or institution. \\
    {Neutral:} Lacks strong emotional cues, purely factual, informational or descriptive. \\[5pt]

    {[Example:]} \\
    Input: {"The wildfire destroyed thousands of homes, leaving families devastated."} \\
    \texttt{<think>} \\
        Explanation: \\
        - The sentence describes destruction and suffering, strongly aligning with {Sadness (0.8)}. \\
        - There is minor frustration in the situation, leading to {Anger (0.05)}. \\
        - The threat aspect contributes to {Fear (0.05)}. \\
        - It lacks positive emotion, optimism, or trust. \\
        - Since the sentence is descriptive but emotionally charged, {Neutral (0.1)} remains minimal. \\
    \texttt{</think>} \\

    \texttt{<output>} \\
        Final Output: \\
        - {Sadness:} 0.8 \\
        - {Anger:} 0.05 \\
        - {Fear:} 0.05 \\
        - {Joy:} 0 \\
        - {Optimism:} 0 \\
        - {Trust:} 0 \\
        - {Neutral:} 0.1 \\
    \texttt{</output>} \\[5pt]
    \hline
    \end{tabular}
    }
    % \end{tcolorbox}
    \label{fig:prompt2}
\end{figure}
\begin{figure}[ht!]
    \centering
    % \small
    \caption{One-shot prompt for topic/subtopic/keyword labeling.}
    % \begin{tcolorbox}[colframe=black, colback=white, boxrule=0.8pt, width=\textwidth]
    \resizebox{1.0\textwidth}{!}{
    \begin{tabular}{|p{22cm}|}
    \hline
    {[Task Description:]} \\
    Given the sentence: "\texttt{<sentence>}",  determine the most relevant topic, subtopic, and keywords based on the definitions below.
    Assign a probability score between 0 and 1 to each possible category under Topic and Subtopic (strictly numeric values only), ensuring the final probability sum is exactly 1 \\

    {[Topic and Subtopic Definitions]} \\
    {Vulnerabilities:} Identifies risk factors that make communities more susceptible to damage. \\
    - {Environmental Vulnerability:} Fragile ecosystems that increase disaster impact. \\
    - {Infrastructure Vulnerability:} Structural weaknesses leading to heightened risk. \\
    - {Economic Vulnerability:} Financial instability due to disasters. \\[5pt]

    {Impact:} Direct effects of extreme weather. \\
    - {Deaths:} Fatalities resulting from disasters. \\
    - {Infrastructure Damage:} Physical destruction of buildings and roads. \\
    - {Economic Damage:} Financial loss due to extreme weather. \\
    - {Homeless:} Displacement due to destruction. \\[5pt]

    {Emergency Response:} Immediate actions taken to mitigate disasters. \\
    - {Evacuation:} Organized movement of people to safety. \\
    - {Community Support:} Aid provided by local and national organizations. \\
    - {Emergency Services:} Rescue and medical response efforts. \\
    - {Communication Strategies:} Plans for disseminating critical information. \\
    {[Example:]} \\
    Input: {"Hurricane Maria devastated Puerto Rico, leaving thousands homeless and without power."} \\
    \texttt{<think>} \\
        Explanation: \\
        - The sentence describes significant destruction, categorizing it under {Impact (0.8)}. \\
        - "Leaving thousands homeless" aligns with {Homeless (0.5)}, and "without power" suggests {Infrastructure Damage (0.3)}. \\
        - Emergency efforts likely followed, so {Emergency Response (0.2)} is relevant, particularly {Emergency Services (0.2)}. \\
    \texttt{</think>} \\
    \texttt{<output>} \\
        Final Output: \\
        - {Topic:} Impact (0.8), Emergency Response (0.2) \\
        - {Sub-Topic:} Homeless (0.5), Infrastructure Damage (0.3), Emergency Services (0.2) \\
        - {Keywords:} devastation, homeless, power outage, Hurricane Maria \\
    \texttt{</output>} \\[5pt]
        - Keywords should not include specific locations. \\
        \hline
    \end{tabular}}
    % \end{tcolorbox}
    \label{fig:prompt3}
\end{figure}

\newpage
\subsection{Annotation Guidelines}
\label{sec:annot_guide}
This section outlines the detailed annotation process for dataset and gold testset preparation.

\subsubsection{Dataset Preparation}
\label{sec:annot_guide_data}

To construct a high-quality alignment dataset tailored to extreme weather event analysis, we begin with the \texttt{ExtremeWeatherNews} corpus, which contains sentence-level reports from various news sources. The goal of this stage is to ensure that only clear, relevant, and logically consistent samples are used for training and evaluation.

\paragraph{Relevance and Clarity Assessment.}
Annotators are first asked to rate each sentence on two dimensions: {clarity} and {relevance} to the context of extreme weather events. 
\begin{itemize}
    \item {Clarity} captures whether the sentence is grammatically sound, unambiguous, and interpretable without requiring additional context.
    \item {Relevance} evaluates whether the sentence pertains to extreme weather events, or if it diverges from this core scope.
\end{itemize}
Sentences rated as \textit{low relevance} or \textit{unclear} are discarded from the dataset to reduce noise during training.

\paragraph{Explanation Verification.}
Following sentence-level filtering, we construct the {ExtremeAlign} dataset, a alignment data contructed by adding task-specific prompts using the sentences in  {ExtremeWeatherNews} dataset aimed at training models for reaosning-aware alignment. Each sample in {ExtremeAlign} includes a sentence, a task-specific output, and an explanation chain justifying the output assignment.

To ensure consistency and interpretability, annotators are instructed to:
\begin{enumerate}
    \item Read both the sentence and its associated explanation.
    \item Verify whether the explanation logically and accurately supports the given task-specific output.
    \item Remove or revise samples where the explanation is incomplete, inconsistent, overly generic, or incorrect based on domain definitions. In particular, explanations must align strictly with the formal definitions of categories. 
\end{enumerate}

\subsubsection{Gold TestSet Annotation}
\label{sec:annot_guide_gold}
To ensure high-quality and consistent annotations for the gold-standard evaluation set, we developed detailed guidelines covering each task in the ExtremeAlign framework: (1) Vulnerability/Impact/Emergency Assessment, (2) Topic/Subtopic Labeling and Keyword Extraction, and (3) Emotion Analysis. These guidelines were designed to balance both explicit task definitions and the flexibility needed to incorporate human reasoning and domain knowledge.
\begin{enumerate}
\item {General Instructions:} Annotators were provided with a set of instructions for each task, including definitions, category descriptions, and examples of edge cases. To minimize ambiguity, we included both inclusion and exclusion criteria for each label, as well as clarifications for frequently confused categories. An initial training round was conducted, followed by calibration sessions to align interpretations across annotators. We also ask them to utilize the general domain knowledge and commonsense reasoning relevant to extreme weather events while annotating.
\item{Vulnerability/Impact/Emergency Assessment:}
Annotators were instructed to assess whether a sentence describes a situation involving vulnerability (pre-existing conditions that increase risk), impact ( measurable damage or consequences of an event), emergency (urgent calls for help or crisis situations). Overlapping categories were allowed, and annotators were asked to assign probability scores across the categories such that they sum to 1. Reasoning for each choice was recorded in free-text format to capture implicit justifications.
\item{Topic/Subtopic Labeling and Keyword Extraction:}
Each sentence was evaluated to determine its thematic focus based on a predefined taxonomy (Tab.~\ref{tab:climate_definitions}). Annotators could select multiple relevant topics and subtopics, and assign confidence scores to reflect relative importance. Annotators were encouraged to extract salient keywords that captured the essence of the sentence.
\item{Emotion Analysis:}
Annotators labeled each sentence with one or more emotions from a defined set (Sadness, Anger, Fear, Joy, Optimism, Trust, Neutral). Probability distributions over the emotion labels were allowed to reflect emotional ambiguity.
\item{Quality Control:}
Annotations were reviewed by a second annotator, and disagreements were adjudicated by a third expert.
\end{enumerate}

\paragraph{Annotator Details and annotator agreement}
Our annotators and experts comprise postdoctoral researchers and senior Ph.D. students with domain expertise in both natural language processing (NLP) and extreme weather analysis. Their interdisciplinary background ensures a nuanced understanding of the linguistic and scientific aspects of the task. Each annotation is evaluated against a well-defined rubric to maintain consistency and reliability across the dataset. Verifiers conduct an additional round of quality checks, resolving ambiguities and refining annotations as needed to meet gold-standard quality. Inter-annotator agreement was measured using Fleiss Kappa score. We obtained an average score of 0.88 for various task showing high agreement between the annotators. Only samples with high agreement were retained in the final gold set.
    
\newpage
\section{ClimaEmpact Online Dashboard}
\label{sec:online_dash}
\begin{figure}[ht!]
    \centering
    \includegraphics[width=\linewidth]{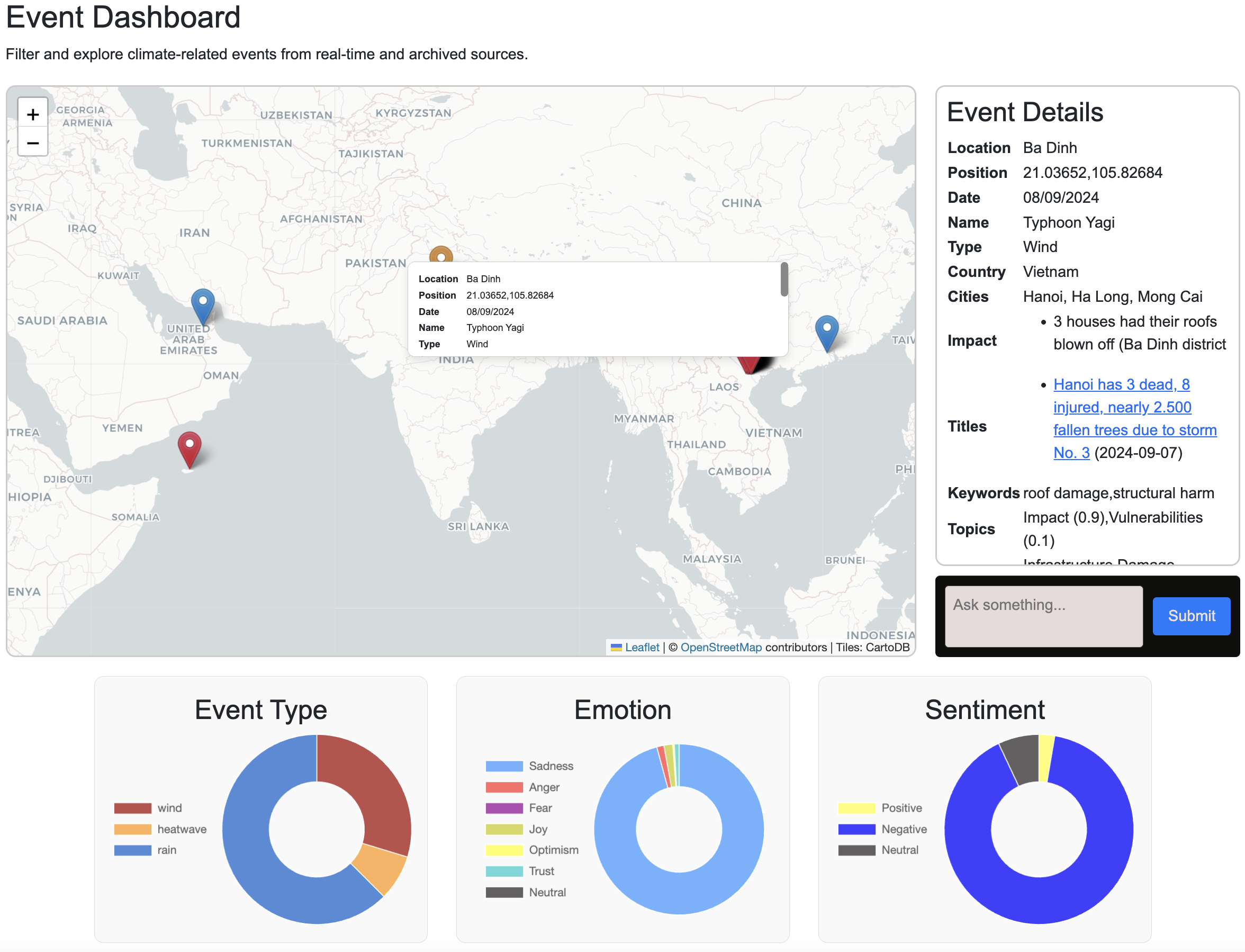}
    \caption{ClimaEmpact Online Dashboard for Extreme Weather Analysis}
    \label{fig:website}
\end{figure}

The ClimaEmpact framework is accessible online, providing rapid and near real-time analysis of extreme weather events. The displayed snapshot showcases various locations impacted by Typhoon Yagi, along with associated impacts, vulnerabilities, topics, subtopics, keywords, and emotion analysis. Fig.~\ref{fig:website} shows the snapshot of our online dashboard.

\end{document}